\documentclass[runningheads]{llncs}

\usepackage[T1]{fontenc}

\usepackage{graphicx}
\usepackage{amsmath}
\usepackage{multirow}
\usepackage{booktabs}
\usepackage{amssymb}
\usepackage{url}
\usepackage{pifont}
\usepackage{color}
\usepackage[misc]{ifsym}

\begin{document}
\title{CT-MVSNet: Efficient Multi-View Stereo with Cross-scale Transformer}


\author{Sicheng Wang \and
Hao Jiang$^{(\textrm{\Letter})}$ \and
Lei Xiang}
\institute{Nanjing University of Information Science and Technology, Nanjing, China \\
\email{\{scwang,jianghao\}@nuist.edu.cn, xl294487391@gmail.com}\\}

\maketitle              
\begin{abstract}
Recent deep multi-view stereo (MVS) methods have widely incorporated transformers into cascade network for high-resolution depth estimation, achieving impressive results. However, existing transformer-based methods are constrained by their computational costs, preventing their extension to finer stages. In this paper, we propose a novel cross-scale transformer (CT) that processes feature representations at different stages without additional computation. Specifically, we introduce an adaptive matching-aware transformer (AMT) that employs different interactive attention combinations at multiple scales. This combined strategy enables our network to capture intra-image context information and enhance inter-image feature relationships. Besides, we present a dual-feature guided aggregation (DFGA) that embeds the coarse global semantic information into the finer cost volume construction to further strengthen global and local feature awareness. Meanwhile, we design a feature metric loss (FM Loss) that evaluates the feature bias before and after transformation to reduce the impact of feature mismatch on depth estimation. Extensive experiments on DTU dataset and Tanks and Temples (T\&T) benchmark demonstrate that our method achieves state-of-the-art results. Code is available at https://github.com/wscstrive/CT-MVSNet.
\end{abstract}
\section{Introduction}

Multi-View Stereo (MVS) is a fundamental task in computer vision that aims to reconstruct the dense geometry of the scene from a series of calibrated images, and has been extensively studied by traditional methods~\cite{patchmatch,gipuma,colmap,furu} for years. Compared to traditional techniques, learning-based MVS methods~\cite{casmvsnet,mvsnet,rmvsnet} use convolutional neural networks for dense depth prediction, which achieve superior advancements in challenging scenarios (e.g., illumination changes, non-Lambertian surfaces, and textureless areas).

Recently, transformer~\cite{multihead} has shown great potential in various tasks~\cite{transmvsnet,wtmvsnet,superglue,loftr,matchformer,mvster}, especially in MVS. By treating MVS as a feature matching task between different images, i.e., finding the pixel with the smallest matching cost under differentiable homography. The transformer-based method~\cite{transmvsnet} can capture long-range context information within and across different images to improve reconstruction quality. Additionally, some networks~\cite{wtmvsnet,mvster,mvs2d} use dimensionality reduction or epipolar constraints to decrease the parameters of the attention mechanism, resulting in reducing the computational burden. However, an interesting phenomenon was observed: these networks prefer to add attention modules to the coarse stage, while the finer stage is just a simple baseline. It makes the whole network appear top-heavy and quite unstable. Thus, how to allocate attention mechanisms in different stages of the feature pyramid network (FPN)~\cite{fpn} without excessive burdens is particularly important.

To address this problem, we revisit the hierarchical design of the FPN and discover that this design has unique characteristics at each stage. Consequently, customizing attention modules based on these individual characteristics becomes a more logical design strategy. In this paper, we propose a novel cross-scale transformer-based multi-view stereo network termed CT-MVSNet. Specifically, we introduce an adaptive matching-aware transformer (AMT) that rearranges and combines interleaved intra-attention and inter-attention blocks to capture intra-image context information and inter-image feature relationships at each stage. For the coarser stage characterized by inherently lower image resolutions, we emphasize using more inter-attention blocks to discern local details from diverse perspectives. As the network progresses to the finer stage, the semantic information from earlier stages gradually permeates and enriches to the finer stage, rendering more intra-attention alone sufficient for enriching texture and local details. Notably, adaptive sampling is used to fine-tune the feature maps at finer stages. It maintains network efficiency but also leads to the loss of critical information in subsequent processes. Therefore, we present a dual-feature guided aggregation (DFGA), which embeds the coarse global feature into the finer cost volume construction, to guide our network to construct cost volume with global semantic awareness and local geometric details. This stratified design avoids imposing additional computational load on the network and ensures that every stage benefits from a mixture of local and global information. Also, our model maintains coarse-stage semantics and fine-granular details, which collectively enhances its stability and efficiency.

In addition, our AMT treats each pixel uniformly, resulting in that a depth map often appears overly smoothed. For this situation, we design a feature metric loss (FM Loss), which weakens discrepancies from feature matching across multiple views while considering geometric consistency and the stability of features post-transformations. Extensive experiments show that our network contributes to enhancing the overall reconstruction quality and achieves state-of-the-art performance on both DTU~\cite{dtu} dataset and
Tanks and Temples~\cite{tnt} benchmark.

In summary, our method presents the following key contributions:

1) We introduce an adaptive matching-aware transformer (AMT) that effectively employs different attention combinations to capture intra-image context information and inter-image feature relationships at each stage of FPN.

2) We propose a dual-feature guided aggregation (DFGA) that aims to enhance the stability of cost volume construction by leveraging the feature information derived from different stages. 

3) We design a novel feature metric loss (FM Loss) that penalizes errors in feature matching across different views to capture subtle details and minor variations within the reconstructed scene.

4) Our method outperforms existing methods, achieving state-of-the-art results on several benchmark datasets, including the DTU dataset and both intermediate and advanced sets of Tanks and Temples benchmark.
\section{Related Works}
\subsection{Learning-based MVS Methods}
Over the past few decades, learning-based MVS has aimed to improve the accuracy and completeness of reconstructed scenes, which has been extensively researched. MVSNet~\cite{mvsnet}, as a pioneering work, roughly divides learning-based MVS into four steps: image feature extraction using 2D CNN, variance-based cost fusion via homography warping, cost regularization through 3D U-Net, and depth regression. Among these steps, the use of 3D U-Net for processing cost volumes introduces significant computational and memory overhead to the entire network, prompting subsequent research to explore alternative methods. Recurrent MVSNet~\cite{aarmvsnet,rmvsnet} leverage recurrent networks to regularize the cost volume along the depth dimension. Cascade MVSNet~\cite{ucsnet,casmvsnet,cvpmvsnet} employ a multi-stage strategy to refine the cost volume by narrowing hypothesis ranges, and this strategy is widely adopted by transformer-based methods.

\subsection{Attention-based Transformer for Feature Matching}
Transformer architecture~\cite{multihead}, initially renowned for its performance in natural language processing (NLP) tasks, has recently garnered considerable attention in the computer vision (CV) community~\cite{vit,linearattention,swintrans}, particularly in feature-matching tasks. Some works~\cite{vector,superglue,loftr,matchformer} have demonstrated the importance of the transformer in capturing long-range global context by using self-attention and cross-attention blocks for feature matching.

TransMVSNet~\cite{transmvsnet} first introduces transformers into MVS, which leverages transformers to capture long-range context information within and across images at the coarsest stage. MVSTER~\cite{mvster} confines attention-based matching to the epipolar lines of the source views, effectively reducing matching redundancy, but it is sensitive to changes in the viewing angle and exhibits unstable matching results. Unlike these methods, we propose CT-MVSNet, which enables efficient and effective feature matching at multiple stages. Compared to TransMVSNet, our network extends long-range context aggregation and global feature interaction to higher-resolution reconstructions. Compared to MVSTER, our network is more robust to inaccurate camera calibrations and poses.

\begin{figure}[t]
\includegraphics[width=\linewidth]{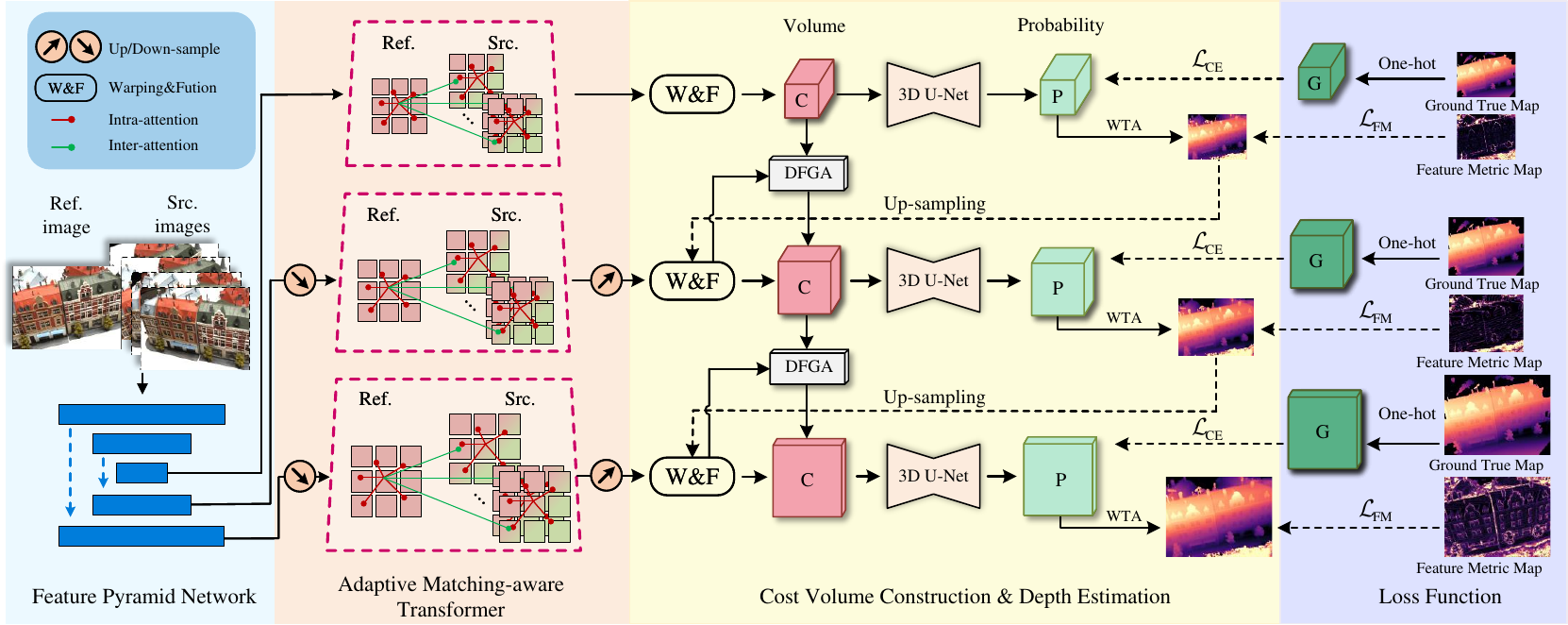}
\caption{\textbf{Illustration of our CT-MVSNet.} FPN first extracts multi-scale features and then introduces AMT, which employs different attention combinations to enhance features at each stage (Sect.~\ref{sec3.2}). W\&F is used to construct cost volume under depth hypotheses and the DFGA is added in finer stages to further refine cost volume construction (Sect.~\ref{sec3.3}). The 3D U-Net is used to obtain a probability volume, and then $winner-take-all$ (WTA) is used to get a depth map (Sect.~\ref{sec3.3}). The CE Loss with one-hot labels is applied to probability volume and the FM Loss with feature metric is applied to the depth map (Sect.~\ref{sec3.4}). Then we update the depth hypotheses by up-sampling to finer stages and finally estimate coarse-to-fine depth maps.}\label{fig:network}
\end{figure}

\section{Methodology}
This section introduces the main process of our network in detail. Firstly, we review the overall architecture of CT-MVSNet in Sect.~\ref{sec3.1}, and then describe in turn the three main contributions: adaptive matching-aware transformer (AMT) in Sect.~\ref{sec3.2}, dual-feature guided aggregation (DFGA) in Sect.~\ref{sec3.3} and feature metric loss (FM Loss) in Sect.~\ref{sec3.4}.

\subsection{Network Overview}\label{sec3.1}
Given a reference image $\mathbf{I}_0\in\bbbr^{3\times H\times W}$ and its neighboring $N-1$ source images $\left\{\mathbf{I}_i\right\}_{i=1}^{N-1}$, as well as their corresponding camera intrinsics $\left\{\mathbf{K}_i\right\}_{i=0}^{N-1}$ and extrinsics $\left\{\mathbf{T}_i\right\}_{i=0}^{N-1}$, CT-MVSNet aims to estimate the reference depth map $\mathbf{D}\in \bbbr^{H'\times W'}$, which aligns $\mathbf{I}_i$ with $\mathbf{I}_0$ in depth ranges $[d_{min},d_{max}]$, to reconstruct a dense 3D point cloud, where $H'$, $W'$ denote the height and width at current stage.

The overall architecture of our network is shown in Fig.~\ref{fig:network}. Follow the coarse-to-fine manner, we first extract multi-scale feature maps $\big\{\mathbf{F}_{i}\in \bbbr^{C'\times H'\times W'}\big\}_{i=0}^{N-1}$ from FPN at resolutions of 1/4, 1/2, and full image scale. To expand the capability for long-range context aggregation and global feature interaction across scales, we introduce the AMT, which employs combinations of interleaved inter-attention and intra-attention blocks within and across multi-view images at each stage. Afterward, we warp 2D transformed features into the 3D frustum of the reference camera using differentiable homography, resulting in feature volumes $\big\{\mathbf{V}_{i}\in \bbbr^{C'\times M'\times H'\times W'}\big\}_{i=0}^{N-1}$, where $C'$, $M'$ denote the channels and depth hypotheses at current stage. These volumes are then fused through a variance-like approach to construct a cost volume $\mathbf{C}\in \bbbr^{M'\times H'\times W'}$. To further enhance cost volume, we introduce DFGA, which embeds the global semantic information and the local texture details into cost volume. Also, we apply a 3D U-Net regularization to cost volume $\mathbf{C}$ to produce a probability volume $\mathbf{P}\in \bbbr^{M'\times H'\times W'}$, which represents the likelihood of the depth value being situated at each depth hypotheses. Finally, we employ the cross-entropy loss to supervise the probability volume and design a FM Loss to ensure precise feature matching across different views.

\subsection{Adaptive Matching-Aware Transformer}\label{sec3.2}
Considering the high memory and computational demands of transformers, most cascade MVS methods that leverage transformers are limited to extracting long-range context information at the coarsest stage and struggle to extend to finer stages. To solve this problem, we introduce AMT, which uses an ingenious method to dismantle the superimposed attention modules of previous methods and arrange and reorganize them to distribute them to appropriate stages. Next, we introduce the preliminaries of attention and then demonstrate how AMT can be seamlessly integrated at various scales to facilitate long-range aggregation and global interactions within and across views.

\subsubsection{Intra-attention and Inter-attention.} 
Common attention mechanisms~\cite{multihead} group feature maps $\left\{\mathbf{F}\right\}_{i=0}^{N-1}$ into queries $\mathbf{Q}$, keys $\mathbf{K}$ and values $\mathbf{V}$. $\mathbf{Q}$ retrieves relevant information from $\mathbf{V}$ according to the attention weights obtained from the dot product of $\mathbf{Q}$ and $\mathbf{K}$ corresponding to each $\mathbf{V}$. However, we follow the linear attention~\cite{linearattention} in place of the common attention~\cite{multihead}. The linear attention effectively reduces the time complexity from $O(n^2)$ to $O(n)$ by leveraging the associative property of multiplication. The kernel function of linear attention is:
\begin{equation}
LinearAttention(\mathbf{Q,K,V}) = \mathrm{\Phi}(\mathbf{Q})(\mathrm{\Phi}(\mathbf{K}^\top )\mathbf{V})
\label{eq:la}
\end{equation}
where $\mathrm{\Phi}(\cdot) = elu(\cdot )+1$ and $elu(\cdot )$ denotes the activation function of the exponential linear unit.

\begin{figure}[t!]
\includegraphics[width=\linewidth]{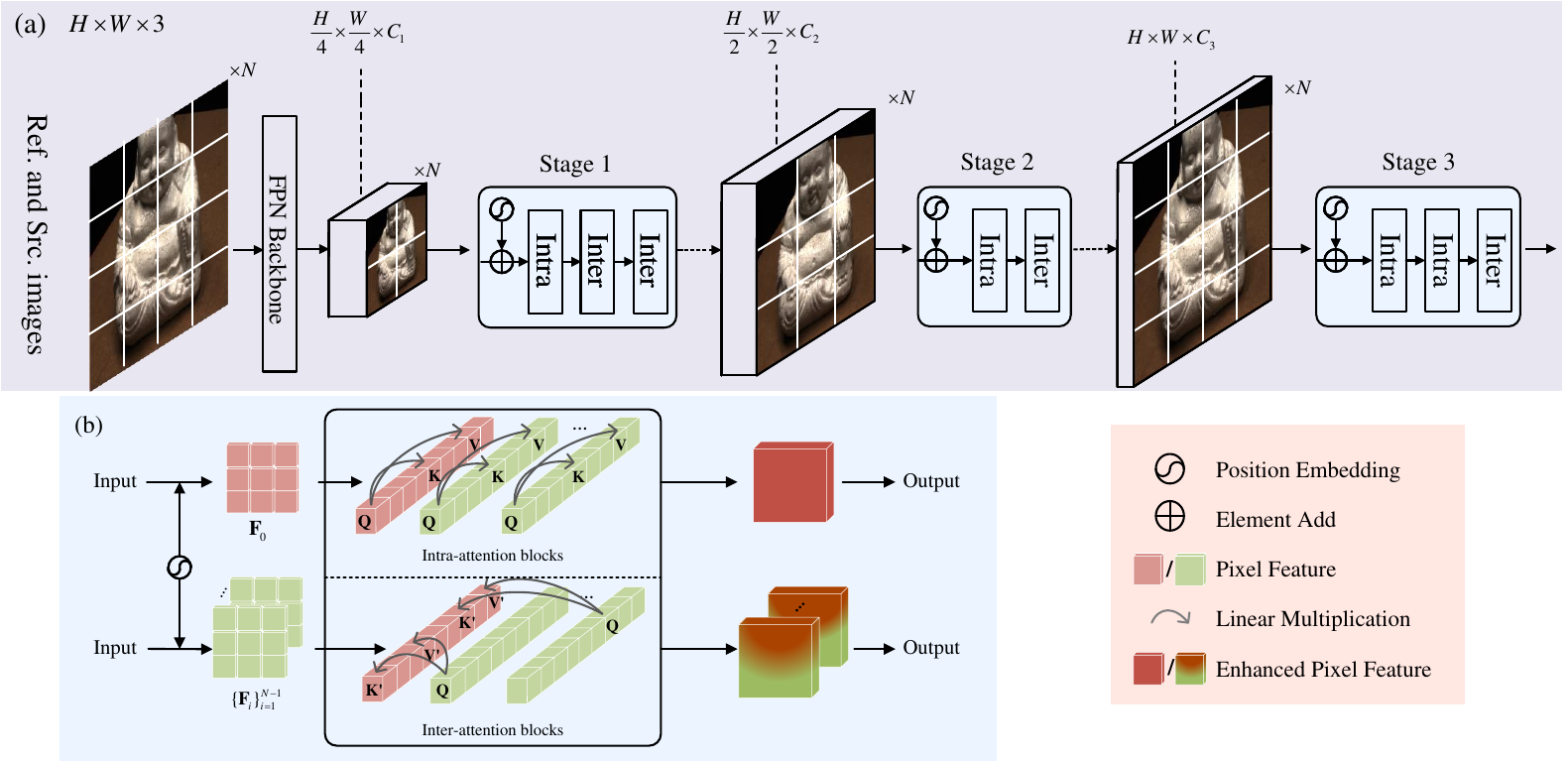}
\caption{\textbf{Illustration of AMT.} (a)Interleaved conbination of intra- and inter-attention at each stage. (b)Internal architecture of two attentions, each stage consist of interleaving-arranged intra-attention (w.r.t. $\mathbf{Q}$, $\mathbf{K}$, $\mathbf{V}$) within images, and inter-attention (w.r.t. $\mathbf{Q}$, $\mathbf{K}'$, $\mathbf{V}'$) across images.}\label{fig:AMT}
\end{figure}

Therefore, we apply linear attention to achieve intra-attention and inter-attention respectively. Fig.~\ref{fig:AMT}(b) shows our attention architectures. For intra-attention, where $\mathbf{Q}$ and $(\mathbf{K}, \mathbf{V})$ originate from the same feature map, it focuses on capturing long-range context information of the given view and emphasizing crucial information for the current task. For inter-attention, where $\mathbf{Q}$ from the source feature maps and ($\mathbf{K}'$, $\mathbf{V}'$) from the reference feature map, it enhances global feature interactions across views and enhance cross-relationships by amalgamating different perspectives. Notably, to prevent the reference feature $\mathbf{F}_0$ from being overly smoothed by inter-attention, this block just enhances the source features $\left\{\mathbf{F}\right\}_{i=1}^{N-1}$.

\subsubsection{AMT Architecture.} 
The overall architecture of our proposed AMT is shown in Fig.~\ref{fig:AMT}(a). To adapt to the feature representation of the FPN at various stages, our AMT utilizes optimal interleaved combinations. These combinations consist of multiple intra-attention and inter-attention blocks tailored for feature maps of different scales. At stage 1, referred to as the coarsest stage, the receptive field for each pixel feature is relatively large, so a singular view cannot offer sufficient information for effective feature enhancement. Thus, we deploy more cross-attention blocks to refer to more different perspective images (i.e., one intra-attention and two inter-attentions). As we transition to the second and third stages, which are characterized by a finer granularity, the resolution noticeably increases. To ensure computational efficiency during AMT processing, we incorporate adaptive sampling, with sampling rates at $1/2$ and $1/4$ respectively. During the second phase, a combination of one intra-attention and inter-attention block serves as a bridge to the most refined phase. Then, at the finest stage, due to a smaller receptive field, the features exhibit rich texture details. Hence, we gravitate towards more intensive intra-attention blocks, involving two intra-attentions and one inter-attention.

\subsection{Cost Volume Construction and Depth Estimation}\label{sec3.3}

Similar to most learning-based MVS methods~\cite{casmvsnet,mvsnet}, for each pixel $\mathbf{p}$ in the reference view, we utilize the differentiable homography to compute the corresponding pixel $\hat{\mathbf{p}}_i$ by warping the source feature $\mathbf{F}(\mathbf{p}_i)$ under the depth hypotheses $\mathit{d}$:

\begin{equation}
\hat{\mathbf{p}}_i = \mathbf{K}_i[\mathbf{R}(\mathbf{K}_{0}^{-1}\mathbf{p}\mathit{d})+\mathbf{t}] 
\label{eq:warp}
\end{equation}
where $\mathbf{R}$ and $\mathbf{t}$ denote the rotation and translation in camera extrinsic $\mathbf{T}$ between reference and source views. $\mathbf{K}_{0}$ and $\mathbf{K}_i$ are the intrinsics of the reference and $i$-th source cameras. To fuse an arbitrary number of input views, a variance-like approach is often adopted to generate the cost volume $\mathbf{C}$, denoted as

\begin{equation}
\mathbf{C}^\ell = \sum_{i=1}^{N-1}\frac{1}{N-1}(\mathbf{V}_i-\mathbf{V}_0)
\end{equation}
where $\mathbf{V}_0$ and $\mathbf{V}_i$ are the warped reference and source feature volumes, $\ell$ represents the layer of feature pyramid network.

\begin{figure}[b!]
\includegraphics[width=\linewidth]{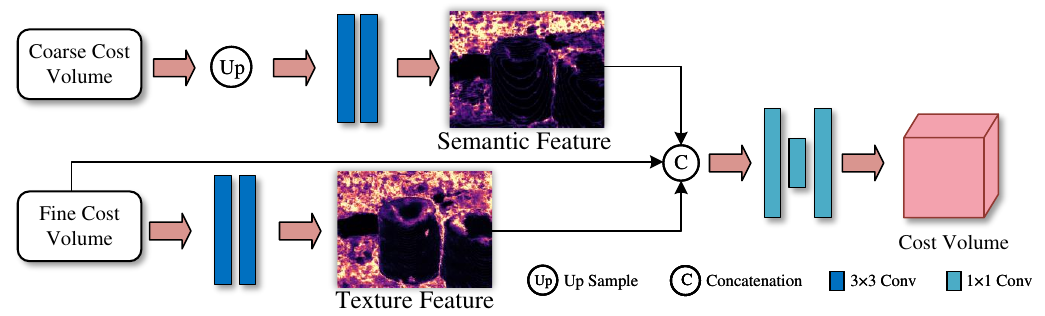}
\caption{\textbf{Illustration of DFGA.} Our DFGA is only applied in the second and third stage.}
\label{fig:dfga}
\end{figure}
\subsubsection{Dual-Feature Guided Aggregation (DFGA).} The details of our proposed DFGA are shown in Fig.~\ref{fig:dfga}. For adaptive sampling at finer stages, the cost volume loses some key feature information. To solve this problem, we propose a DFGA to enrich cost volumes. The method first uses a 3$\times$3 2D convolution operation to extract the coarse-scale global semantic information of the previous stage and the local texture information in the initial cost volume of the current stage. Then, we fuse these two types of information and the initial cost volume through a skip connection strategy to obtain our cost volume. Also, we use a 1$\times$1 2D convolution operation to further refine cost volume. The final cost volume that synthesizes global and local features is computed as follows:

\begin{equation}
\hat{\mathbf{C}}^\ell=DFGA[\mathbf{C}^\ell,\mathbf{C}_\uparrow^{\ell-1}]
\end{equation}

\begin{equation}
DFGA[\;,\;]=\mathrm{\Sigma}_{1\times1}\left[\mathrm{\Sigma}_{3\times 3}(\mathbf{C}_\uparrow^{\ell-1}),\;\mathrm{\Sigma}_{3\times 3}(\mathbf{C}^\ell),\;\mathbf{C}^\ell\right] 
\end{equation}
where $\mathbf{C}_\uparrow^{\ell-1}$ represents an up-sampled cost volume from the previous stage, and $\mathbf{C}^\ell$ is the cost volume at the current stage,  $[\cdot,\cdot]$ is concatenation operation, $\mathrm{\Sigma}_{1\times1}$ and $\mathrm{\Sigma}_{3\times 3}$ represent the convolution of two different receptive fields. The 3$\times$3 receptive field excels at capturing spatially relevant local structures and contextual nuances, while the 1$\times$1 receptive field bolsters feature representation, promotes cross-channel feature interaction, and curtails both parameters and computational complexity.
\subsubsection{Depth Estimation.} We employ a 3D U-Net to regularize the DFGA-enhanced cost volume into a probability volume $\mathbf{P}$. Applying $softmax$ and $winner-take-all$ (WTA) strategy to probability volume yields our final depth map:
\begin{equation}
\mathbf{D} = WTA\left(\underset{d\in\left\{d_m\right\}_{m=1}^M}{softmax}\left(\mathbf{P}^{(d)}\right)\right)
\end{equation}
where $d_m$ refers to the depth hypotheses of the $m$-th level.

\subsection{Loss Functions}\label{sec3.4}
For MVS, depth estimation is often modeled as a probability distribution problem. Thus, pixel-wise classification methods~\cite{transmvsnet,aarmvsnet} are more suitable than L1-based regression~\cite{casmvsnet,mvsnet} in capturing accurate confidence maps. We use cross-entropy loss for pixel-wise supervision:

\begin{equation}
    \mathcal{L}_{\mathrm{CE} } = \sum_{\mathbf{p} \in \mathrm{\Psi} } \sum_{d = d_{1} }^{d_{M}}-\mathbf{G}^{(d)}(\mathbf{p})\log{[\mathbf{P}^{(d)}(\mathbf{p})]} 
  \label{eq:ce}
\end{equation}
where $\mathbf{G}^{(d)}(\mathbf{p})$ denote ground truth probability of depth hypotheses $d$ at pixel $\mathbf{p}$. $\mathrm{\Psi}$ is the set of valid pixels with ground truth precision.

Moreover, we design a novel FM Loss to explore geometric and feature consistency between reference view and source views, which achieves more accurate depth estimation by supervising depth maps and penalizing errors in feature matching across different views. Using differentiable homography, we project pixel $\mathbf{p}$ from the reference image onto the $i$-th source image, resulting in $\mathbf{p}_i'$, and then warp back to the reference image, yielding $\mathbf{p}''$:
\begin{equation}
\left\{\begin{array}{l}
  \mathbf{p}_i' = \mathbf{T}_0^{-1} \mathbf{K}_0^{-1}\mathbf{D}_0(\mathbf{p})\mathbf{p}\\[0.6em]
  \mathbf{p}''=(\mathbf{K}_i \mathbf{T}_i \mathbf{p}')/\mathbf{D}_i^{gt}(\mathbf{p}_i')
\end{array}\right.
\label{eq:reproj}
\end{equation}
where $\mathbf{D}_i^{gt}(\mathbf{p}_i')$ denotes the ground truth depth value of $\mathbf{p}_i'$. FM Loss evaluates the consistency of each feature across views to ensure their stability and reliability under varying conditions. Our final FM Loss is:

\begin{equation}
\mathcal{L}_{\mathrm{FM}} = (\mathrm{\Upsilon})(\left|\left|\mathbf{D}_0(\mathbf{p})- \mathbf{D}_0^{gt}(\mathbf{p})\right|\right|_2)
\label{eq:fm3}
\end{equation}
where $\mathbf{D}_0^{gt}(\mathbf{p})$ represents the depth value predicted by our network at pixel $\mathbf{p}$. To take into account in depth the differences between original features and their mappings after various transformations, we add feature metric weights $\mathrm{\Upsilon}$, which is denoted as:
\begin{equation}
\mathrm{\Upsilon} =\underset{\upsilon\in[\upsilon_1,\upsilon_2]}{log} \left | \upsilon \right|, \;\ \upsilon  = \frac{\hat{\mathbf{F}}_0(\mathbf{p}'')-\hat{\mathbf{F}}_0(\mathbf{p})}{\mathbf{F}_0(\mathbf{p}'')-\mathbf{F}_0(\mathbf{p})+\varepsilon}
\label{eq:fm4}
\end{equation}
where $\upsilon$ is feature bias, $\mathbf{F}_0(\mathbf{p}'')$ and $\mathbf{F}_0(\mathbf{p})$ denote the features and back-projected features before transformer, and $\hat{\mathbf{F}}_0(\mathbf{p}'')$ and $\hat{\mathbf{F}}_0(\mathbf{p})$ represent the features and back-projected features after transformer, $\varepsilon$ ensures the denominator does not approach zero, $[\upsilon_1,\upsilon_2]$ defines the acceptable range for feature deviations. 

In summary, the loss function consists of $\mathcal{L}_{\mathrm{CE} }$ and $\mathcal{L}_{\mathrm{FM}}$:
\begin{equation}
    \mathcal{L} = \sum_{\ell=1}^{L}\lambda_{1}^{\ell}\mathcal{L}_{\mathrm{CE}}+ \lambda_{2}^{\ell} \mathcal{L}_{\mathrm{FM}}   
  \label{eq:totalloss}
\end{equation}
where $\lambda_1^*$ = 2, $\lambda_2^*$ = 1.2 among each stage in our experiments, and $L$ is the number of stages.

\section{Experiment}
\subsection{Datasets.} 
\textbf{DTU}~\cite{dtu} dataset is an indoor multi-view stereo dataset with 124 scenes with 49 views and 7 illumination conditions. It contains ground-truth point clouds captured under well-controlled laboratory conditions for evaluation. \textbf{Tanks and Temples(T\&T)}~\cite{tnt} dataset is a large-scale dataset captured in realistic indoor and outdoor scenarios, which contains an intermediate subset of 8 scenes and an advanced subset of 6 scenes. \textbf{BlendedMVS}~\cite{blendedmvs} dataset is a large-scale synthetic MVS training dataset containing cities, sculptures, and small objects, split into 106 training scans and 7 validation scans.
\subsection{Implementation Details.}
We implemented CT-MVSNet with PyTorch, which is trained and evaluated on DTU dataset~\cite{dtu}. Also, BlendedMVS dataset~\cite{blendedmvs} is used to fine-tune our model for evaluation on T\&T dataset~\cite{tnt}. For the coarse-to-fine network, we use $L=3$ layer pyramids, the layer number of depth hypotheses $M$ is set to 48, 32, 8 for each level, and the depth sampling range is set to $425\mathit{mm} \sim 935\mathit{mm}$. We set $\upsilon_1$ and $\upsilon_2$ are 0.6 and 1.7 for acceptable ranges in feature deviations. For training on DTU, the number of input images is set to $N=5$ with a resolution of $512\times640$. As for fine-tuning on BlendedMVS, we set $N=5$ with $576\times768$ image resolutions. We used the Adam optimizer with a learning rate of 0.001 and trained the model for 16 epochs, reducing the learning rate by 0.5 after 6, 8, and 12 epochs. We trained the model on 8 NVIDIA GeForce RTX 2080Ti GPUs, with a batch size of 1. The training of CT-MVSNet typically takes around 18 hours and consumes approximately 11\textit{GB} of memory on each GPU.

\begin{figure}[t!]
\includegraphics[width=\textwidth]{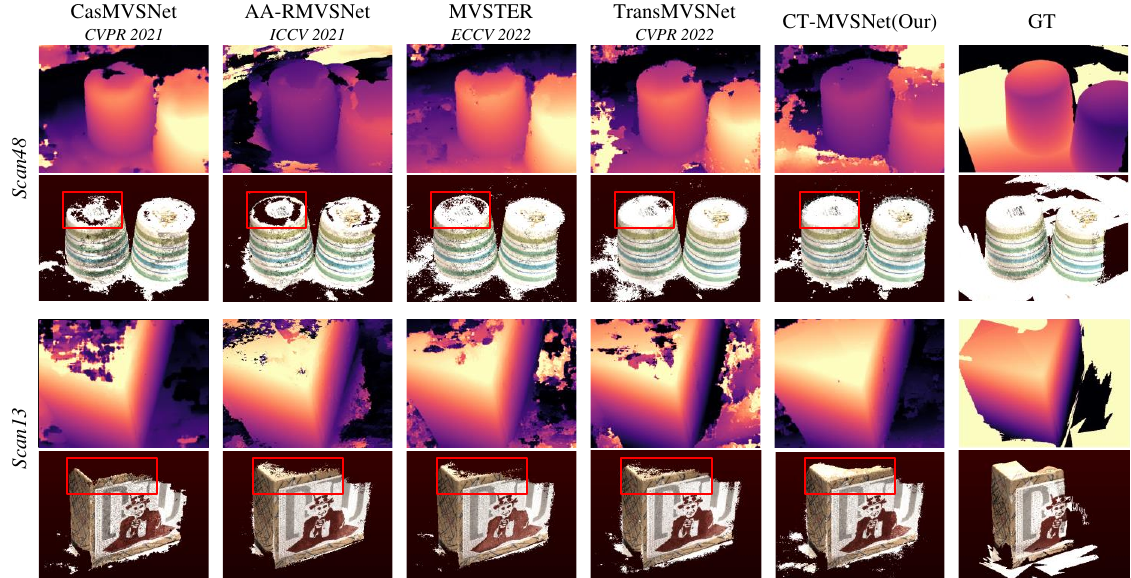}
\caption{\textbf{Point clouds error comparison of state-of-the-art methods on DTU dateset.} The first and third rows are estimated depth maps while others are point cloud reconstruction results.} 
\label{fig:dtu}
\end{figure}
\begin{table}[t!]
\setlength\tabcolsep{3pt}
  \centering
  \caption{\textbf{Quantitative comparison on the DTU dataset (lower is better).} The lower is better for Accuracy (Acc.), Completeness (Comp.), and Overall. The best results are in \textbf{bold} and the second best results are in \underline{underlined}.} 
  \label{tab:dtu}
  \begin{tabular}{l|c|c|c|c}
    \hline
    \textbf{Method} & \;\textbf{Years}\; & \textbf{Acc.(\textit{mm})} & \textbf{Comp.(\textit{mm})} & \textbf{Overall$\downarrow$(\textit{mm})}\\
    \hline
    Gipuma~\cite{gipuma} & ICCV 2015 & \textbf{0.283} & 0.873 & 0.578\\
    COLMAP~\cite{colmap} & CVPR 2016 & 0.400 & 0.664 & 0.532\\
    \hline
    R-MVSNet~\cite{rmvsnet} & CVPR 2019 & 0.385 & 0.459 & 0.422\\
    CasMVSNet~\cite{casmvsnet} & CVPR 2020 & 0.325 & 0.385 & 0.355\\
    CVP-MVSNet~\cite{cvpmvsnet} & CVPR 2020 & \underline{0.296} & 0.406 & 0.351\\
    \hline
    AA-RMVSNet~\cite{aarmvsnet} & ICCV 2021 & 0.376 & 0.339 & 0.357\\
    EPP-MVSNet~\cite{eppmvsnet}& CVPR 2021 & 0.413 & 0.296 & 0.355\\
    Patchmatchnet~\cite{patchmatchnet} & CVPR 2021 & 0.427 & 0.277 & 0.352\\
    RayMVSNet~\cite{raymvsnet} & CVPR 2022 & 0.341 & 0.319 & 0.330\\
    Effi-MVSNet~\cite{effimvs} & CVPR 2022 & 0.321 & 0.313 & 0.317\\
    MVS2D~\cite{mvs2d} & CVPR 2022 & 0.394 & 0.290 & 0.342\\
    TransMVSNet~\cite{transmvsnet} & CVPR 2022 & 0.321 & 0.289 & \underline{0.305}\\
    UniMVSNet~\cite{unimvsnet} & CVPR 2022 & 0.352 & 0.278 & 0.315\\
    MVSTER~\cite{mvster} & ECCV 2022 & 0.350 & \underline{0.276} & 0.313\\
    \hline
    CT-MVSNet (Ours) & MMM 2024 & 0.341 & \textbf{0.264} & \textbf{0.302}\\
    \hline
  \end{tabular}
\end{table}

\begin{table}[t!]
  \centering
  \caption{\textbf{Quantitative results on the Tanks and Temples dataset (higher is better).} \textbf{Bold} represents the best while \underline{underlined} represents the second-best.}
  \label{tab:tnt}
  \resizebox{\linewidth}{!}{
  \begin{tabular}{c|ccccccccc|ccccccc}
     \hline
     \multirow{2}{*}{\textbf{Method}} & \multicolumn{9}{c|}{\textbf{Intermediate}} & \multicolumn{7}{c}{\textbf{Advanced}} \\
     \cline{2-17}
     & \textbf{Mean}$\uparrow$ & Family & Francis & Horse & L.H. & M60 & Panther & P.G. & Train & \textbf{Mean}$\uparrow$ & Auditorium & Ballroom  & Courtroom&Museum & Palace & Temple\\
     \hline
     COLMAP \cite{colmap} & 42.14 & 50.41 & 22.25 & 26.63 & 56.43 & 44.83 & 46.97 & 48.53 & 42.04 & 27.24 & 16.02 & 25.23 & 34.70 & 41.51 & 18.05 & 27.94 \\
     R-MVSNet \cite{rmvsnet} & 50.55 & 73.01 & 54.46 & 43.42 & 43.88 & 46.80 & 46.69 & 50.87 & 45.25 & 29.55 & 19.49 & 31.45 & 29.99 & 42.31 & 22.94 & 31.10 \\
     CasMVSNet \cite{casmvsnet} & 56.42 & 76.36 & 58.45 & 46.20 & 55.53 & 56.11 & 54.02 & 58.17 & 46.56 & 31.12 & 19.81 & 38.46 & 29.10 & 43.87 & 27.36 & 28.11 \\
     Patchmatchnet \cite{patchmatchnet} & 53.15 & 66.99& 52.64 & 43.24 & 54.87 & 52.87 & 49.54 & 54.21 & 50.81 & 32.31 & 23.69 & 37.73 & 30.04 & 41.80& 28.31& 32.29\\
     AA-RMVSNet \cite{aarmvsnet} & 61.51 & 77.77 & 59.53 & 51.53 & \textbf{64.02} & \textbf{64.05} & 59.47 & \underline{60.85} & 55.50 & 33.53 & 20.96 & 40.15 & 32.05 & 46.01 & 29.28 & 32.71 \\
     Effi-MVSNet \cite{effimvs} & 56.88 & 72.21 & 51.02 & 51.78 & 58.63 & 58.71 & 56.21 & 57.07 & 49.38 & 34.39 & 20.22 & 42.39 & 33.73 & 45.08 & 29.81 & 35.09 \\
     TransMVSNet \cite{transmvsnet} & \underline{63.52} & \underline{80.92} & \underline{65.83} & \underline{56.89} & 62.54 & 63.06 & \underline{60.00} & 60.20 & \underline{58.67} & 37.00 & 24.84 & \underline{44.59} & 34.77 & 46.49 & \underline{34.69} & 36.62 \\
     GBi-Net \cite{gbinet} & 61.42 & 79.77 & \textbf{67.69} & 51.81 & 61.25 & 60.37 & 55.87 & 60.67 & 53.89 & \underline{37.32} & \textbf{29.77} & 42.12 & \textbf{36.30} & \underline{47.69} & 31.11 & 36.93 \\ 
     MVSTER \cite{mvster} &60.92 & 80.21 & 63.51 & 52.30 & 61.38 & 61.47 & 58.16 & 58.98 & 51.38 & 37.53 & 26.68 & 42.14 & \underline{35.65} & \textbf{49.37} & 32.16 & \textbf{39.19} \\
     \hline
     CT-MVSNet (Ours) & \textbf{64.28} & \textbf{81.20} & 65.09 & \textbf{56.95} & \underline{62.60} & \underline{63.07} & \textbf{64.83} & \textbf{61.82} & \textbf{58.68} & \textbf{38.03} & \underline{28.37} & \textbf{44.61} & 34.83 & 46.51 & \textbf{34.69} & \underline{39.15} \\
     \hline
  \end{tabular}}
\end{table}

\begin{figure}[t!]
\includegraphics[width=\textwidth]{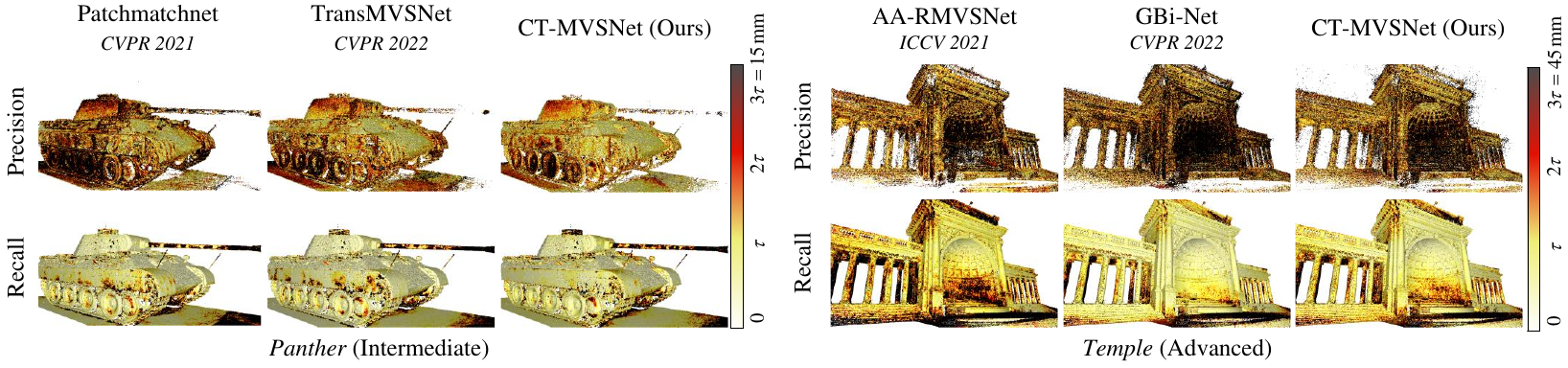}
\caption{\textbf{Point clouds error comparison of state-of-the-art methods on the Tanks and Temples dataset.} $\tau$ is the scene-relevant distance threshold determined officially and darker means larger errors.} \label{fig:tnt}
\end{figure}
\subsection{Benchmark Performance}
\subsubsection{DTU.} We evaluated our proposed method on DTU evaluation set at $864\times1152$ resolution with $N=5$. The visualized results are shown in Fig.~\ref{fig:dtu}, CT-MVSNet produces notably accurate depth estimations and comprehensive point clouds, particularly on low-texture and edge surfaces. Furthermore, the robustness of our method is further demonstrated by its performance on two scenes: $\mathit{scan}13$ and $\mathit{scan}48$. These scenes, which account for illumination changes and reflections, are considered the most challenging within the DTU evaluation set. The quantitative results are summarized in Table~\ref{tab:dtu}. Here, accuracy and completeness are reported using the official MATLAB code. The overall metric represents the average of accuracy and completeness measures. Our approach achieves the highest completeness and overall metric compared to other state-of-the-art methods. 

\subsubsection{Tanks and Temples.} We further validate the generalization capability of our method on the T\&T dataset. Fig.~\ref{fig:tnt} illustrates the error comparison of the reconstructed point clouds on the scene Pather of intermediate and Temple of advanced, and our method exhibits higher precision and recall, especially in sophisticated and low-texture surfaces. The quantitative results on both intermediate and advanced sets are reported in Table~\ref{tab:tnt}, the Mean is the average score of all scenes, CT-MVSNet achieves the highest results in many complex scenes. Besides, there is an official website for online evaluation of T\&T benchmark.

\subsection{Ablation Study}
Table~\ref{tab:modules} shows the ablation results of our CT-MVSNet. Our baseline is CasMVSNet~\cite{casmvsnet}, and all experiments were conducted with the same hyperparameters.

\subsubsection{Effect of AMT.}
As shown in Table~\ref{tab:amt}, we conducted different combinations of attention blocks. By adding a combination of attention blocks at each stage to make the network always focus on global information acquisition, as depicted in Fig.~\ref{fig:dtu}, we have achieved outstanding reconstruction results in texture holes. Also, the self-adaptive sampling strategy guarantees time and memory consumption.
\subsubsection{Effect of DFGA.} 
In Table~\ref{tab:dfga}, we conducted ablation study on semantic feature (SF) and texture feature (TF) to verify effectiveness. DFGA makes cost volume between stages to establish connections, thereby promoting the restoration and supplementation of information after sampling strategy and stage progress.
\subsubsection{Effect of FM Loss.}
We unify the weight $\lambda_1$ of CE Loss to 2 and perform ablation study on the weight $\lambda_2$ of FM Loss in Table~\ref{tab:fml}. Based on feature consistency, network focuses more on vital information, and the combination of AMT brings about feature enhancement and fully plays global feature perception.
\begin{table}[htb]
\begin{minipage}[b]{0.48\linewidth}
  \centering
  \caption{Ablation results with different components on DTU.}
  \label{tab:modules}
  \resizebox{0.9\textwidth}{!}{
  \begin{tabular}{@{}c|cccc|ccc@{}}
    \hline
      & \multicolumn{4}{c|}{\textbf{Module Settings}} & \multicolumn{3}{c}{\textbf{Mean Distance}}\\
     \cline{2-8} 
      & CE Loss & AMT & DFGA & FM Loss & Acc. & Comp. & Overall.\\
    \hline
    (a) & & & & & 0.360 & 0.338 & 0.349\\
    (b) & \checkmark & & & &0.342 & 0.325& 0.333\\
    (c) & \checkmark & \checkmark & & & 0.371 & \textbf{0.262} & 0.316\\
    (d) & \checkmark & \checkmark & \checkmark & & 0.346 & 0.269 & 0.308 \\
    (e) & \checkmark & \checkmark & \checkmark & \checkmark & \textbf{0.341} & 0.264 & \textbf{0.302} \\
    \hline
  \end{tabular}}\medskip
\end{minipage}
\hfill
\begin{minipage}[b]{0.48\linewidth}
  \centering
  \caption{Ablation study of the number of optimization combinations on DTU. }
  \label{tab:amt}
  \resizebox{\textwidth}{!}{
  \begin{tabular}{@{}c|c|c|ccc|cc@{}}
    \bottomrule
      & intra- & inter- & Acc. & Comp. & Overall. & Mem.(\textit{GB}) & Time(\textit{s})\\
    \hline
    (a) & 4,0,0 & 4,0,0 & 0.337 & 0.293 & 0.315 & 3990 & 1.05\\
    (b) & 2,0,0 & 2,0,0 & 0.344 & 0.298 & 0.321 & \textbf{3875} & \textbf{0.88}\\
    (c) & 1,2,0 & 2,1,0 & 0.356 & 0.274 & 0.309 & 4034 & 1.06\\
    (d) & 1,2,2 & 2,2,1 &  0.346 & \textbf{0.262} & 0.304 & 4276 & 1.20\\
    (e) & 1,1,2 & 2,1,1 &  \textbf{0.341} & 0.264 & \textbf{0.302} & 4117 & 1.11\\ 
    \toprule
  \end{tabular}}\medskip
\end{minipage}

\vspace{1em} 

\begin{minipage}[b]{0.48\linewidth}
  \centering
  \caption{Ablation study concerning the SF and TF in the construction process of cost volume.}
  \label{tab:dfga}
  \resizebox{0.7\textwidth}{!}{
  \begin{tabular}{@{}c|c|ccc@{}}
    \hline
      & DFGA &  Acc. & Comp. & Overall. \\
    \hline
    (a) & SF & 0.346 & 0.269 & 0.308 \\
    (b) & SF+TF &  0.344 & 0.266 & 0.305\\
    (c) & Ours & \textbf{0.341} & \textbf{0.264} & \textbf{0.302}\\
    \hline 
  \end{tabular}}\medskip
\end{minipage}
\hfill
\begin{minipage}[b]{0.48\linewidth}
  \centering
  \caption{Ablation study on the different weight parameters $\lambda_1$, $\lambda_2$ of loss function.}
  \label{tab:fml} 
  \resizebox{0.55\linewidth}{!}{
  \begin{tabular}{@{}c|cc|ccc@{}}
    \hline
      & $\lambda_1$ & $\lambda_2$  & Acc. & Comp. & Overall.\\
    \hline
    (a) & 2 & 0 & 0.346 & 0.269 & 0.308 \\
    (b) & 2 & 2 & 0.356 & 0.265 & 0.310 \\
    (c) & 2 & 1.5 &  0.342 & 0.266 & 0.304 \\
    (d) & 2 & 1.2 &\textbf{0.341} & \textbf{0.264}& \textbf{0.302} \\
    (e) & 2 & 1 &  0.345 & 0.274 & 0.309 \\
    \hline 
  \end{tabular}}\medskip
\end{minipage}
\label{fig:res}
\end{table}

\section{Conclusion}
In this paper, we propose a novel learning-based MVS network, named CT-MVSNet, designed for long-range context aggregation and global feature interaction at different scales. Specifically, we introduce an adaptive matching-aware transformer (AMT) that effectively leverages intra- and inter-attention blocks within and across images at each stage. In addition, we present a dual-feature guided aggregation (DFGA) to guide our network to construct cost volume with semantic awareness and geometric details. Meanwhile, we design a feature metric loss (FM Loss) to penalize errors in feature matching across different views. Extensive experiments prove that CT-MVSNet achieves superior performance in 3-D reconstruction. In the future, we intend to expand upon the merits of CT, positioning it to replace the role of FPN as the feature encoder in MVSNet.

\subsubsection{Acknowledgements} This work is supported by the National Natural Science Foundation of China (NSFC) (No. 62101275).

%
%
%
\bibliographystyle{splncs04}
\bibliography{reference}

\end{document}